\documentclass[11pt,twocolumn]{article}

\usepackage[letterpaper,margin=1in]{geometry} 
\usepackage{times}
\usepackage{microtype}
\usepackage{booktabs}
\usepackage{amsmath,amssymb}
\usepackage{graphicx}
\usepackage{xspace}
\usepackage[hidelinks]{hyperref}
\usepackage{enumitem}
\usepackage{siunitx}
\usepackage{multirow}
\usepackage{wrapfig}
\usepackage{listings}
\usepackage{caption}
\usepackage{subcaption}
\usepackage{tikz}
\usepackage{pgfplots}
\usepackage{float}
\usepackage[section]{placeins} 
\usetikzlibrary{positioning,arrows.meta,fit,shadows.blur,calc}
\pgfplotsset{compat=1.18}
\usepackage{multicol}
\usepackage{array} 

\usepackage{dblfloatfix} 
\newcommand{\benchmark}{Falcon\xspace}
\newcommand{\exacc}{Exact Result Accuracy\xspace}

\title{Falcon: A Comprehensive Chinese Text-to-SQL Benchmark for Enterprise-Grade Evaluation}

\author{
  Wenzhen Luo \quad Wei Guan \quad Yifan Yao \quad Yimin Pan \quad Feng Wang \quad Zhipeng Yu  \\
  \quad Zhe Wen \quad Liang Chen \quad Yihong Zhuang \\
  Ant Group \\
  \texttt{https://github.com/eosphoros-ai/Falcon}
}
\date{} 

\begin{document}
\maketitle

\begin{abstract}
We introduce \benchmark, a cross-domain Chinese text-to-SQL benchmark grounded in an enterprise-compatible dialect (MaxCompute/Hive). It contains 600 Chinese questions over 28 databases; 77\% require multi-table reasoning and over half touch more than four tables. Each example is annotated along SQL-computation features and Chinese semantics. For evaluation, we release a robust execution comparator and an automated evaluation pipeline, under which all current state-of-the-art large-scale models (including Deepseek) achieve accuracies of at most 50\%. Major errors originate from two sources: (1) schema linking in large enterprise landscapes — hundreds of tables, denormalized fields, ambiguous column names, implicit foreign-key relations and domain-specific synonyms that make correct join/column selection difficult; and (2) mapping concise, colloquial Chinese into the exact operators and predicates required for analytics — e.g., choosing the correct aggregation and group-by keys, expressing time windows and granularities, applying unit conversions, handling NULLs and data-quality rules, and formulating nested or windowed subqueries. \benchmark therefore targets Chinese-specific semantics and enterprise dialects (abbreviations, business jargon, fuzzy entity references) and provides a reproducible middle ground before full production deployment by using realistic enterprise schemas, query templates, an execution comparator, and an automated evaluation pipeline for end-to-end validation.
\end{abstract}

\section{Introduction}
Natural-language interfaces to databases (Text-to-SQL) aim to make structured data accessible to non-technical analysts by translating human queries into executable SQL. This capability is increasingly critical as organizations seek to democratize data exploration, accelerate decision-making, and reduce the engineering bottleneck for ad-hoc analytics. In enterprise settings, however, production workloads introduce practical constraints — very wide, denormalized schemas, domain-specific naming conventions, large-scale data partitions, dialect-specific SQL features (e.g., MaxCompute/Hive), and strict correctness and performance requirements — that amplify the difficulty of reliable semantic parsing.

These deployment realities interact with language-specific phenomena in ways that standard academic benchmarks do not fully capture. In Chinese enterprise contexts, analysts typically pose concise, elliptical queries using business jargon, mixed-language entity mentions, and implicit temporal or aggregation intents; at the same time schemas are often English-labeled and organized according to historical, product- or team-specific conventions. Together, these factors create a benchmark gap: models that perform well on prior cross-domain datasets may still fail to produce correct, executable SQL in real enterprise workflows.

Foundational cross-domain benchmarks have enabled steady progress: Spider~\cite{yu2019spider-paper} evaluates compositional generalization on unseen schemas; SParC~\cite{sparc-site} introduces multi-turn interactions; BIRD~\cite{bird-paper} emphasizes database values, external knowledge, and efficiency; Spider~2.0~\cite{lei2025spider2-paper} moves to enterprise workflows requiring long-context retrieval, heterogeneous dialects, and multi-query pipelines.

Despite this progress, Chinese text-to-SQL remains under-served, particularly under enterprise dialects (e.g., MaxCompute/Hive) and wide multi-table schemas. Analysts frequently ask Chinese questions over English-labelled schemas with business-style phrasing and ellipsis. Existing Chinese resources such as CSpider~\cite{min2019cspider-paper} are invaluable but largely translation-based and schema-centric; they under-emphasize Chinese-specific semantics and enterprise constraints. At the other extreme, Spider~2.0 requires complex agentic orchestration, raising the barrier for rapid iteration in Chinese and dialect-compatible environments. To address these shortcomings, we introduce Falcon — a benchmark and tooling suite designed to bridge research and deployment by advancing Chinese Text-to-SQL performance in realistic enterprise environments.

\paragraph{Contributions.}
\begin{itemize}[leftmargin=1.2em]
  \item Establish an enterprise-oriented benchmark comprising 600 Chinese questions across 28 databases, where 77\% require multi-table reasoning and over half involve joins spanning more than four tables, with full support for MaxCompute/Hive dialect features. 
  \item Introduce a dual annotation framework that systematically captures both SQL computational patterns (including join topologies, nesting structures, and aggregation usage) and Chinese-specific linguistic phenomena (such as ellipsis resolution, business terminology mapping, and numeric expression variants). This fine-grained annotation enables detailed error analysis and model diagnostics.
  \item Develop a robust evaluation system featuring a schema-aware SQL comparator that handles common syntactic variations (column/table aliases, projection ordering, and equivalent syntax forms) while maintaining strict semantic equivalence.
  \item Provide a systematic evaluation of current state-of-the-art large language models (including Deepseek) on Falcon using our execution comparator and automated pipeline, showing that no tested model exceeds 50\% execution accuracy; we release model outputs, prompts, and evaluation code for reproducibility.
\end{itemize}

\section{Related Work and Existing Datasets}

The evolution of text-to-SQL benchmarks reflects three key dimensions of progress in the field: generalization capability, linguistic complexity, and engineering practicality. Early datasets like ATIS~\cite{atis-1990} and GeoQuery~\cite{geoquery-1996} established basic semantic parsing paradigms but suffered from template memorization due to question-based splits, where identical SQL patterns appeared in both training and test sets. This fundamental limitation persisted until the introduction of database-based splits in Spider~\cite{yu2019spider-paper}, which established the current standard for evaluating cross-domain generalization on completely unseen schemas.

Modern benchmarks have since diverged along two evolutionary paths. One direction, exemplified by SParC~\cite{sparc-site} and DuSQL~\cite{dusql-2020}, extends complexity along the conversational dimension with multi-turn interactions and context-dependent queries. The other direction, represented by BIRD~\cite{bird-paper} and Spider 2.0~\cite{lei2025spider2-paper}, emphasizes real-world engineering constraints including value grounding, execution efficiency, and heterogeneous system integration.

For Chinese language processing specifically, CSpider~\cite{min2019cspider-paper} provided the crucial first step by translating Spider's English questions while preserving the original database schemas. However, our analysis reveals three critical limitations in current Chinese benchmarks: (1) they maintain an artificial separation between Chinese questions and English schema labels, ignoring real-world scenarios where both query and metadata are in Chinese; (2) they underestimate the prevalence of business-specific phrasing and numeric expressions in enterprise queries; and (3) they fail to account for dialect variations in widely-used systems like MaxCompute and Hive.

\benchmark advances the field by simultaneously addressing these gaps while maintaining precise comparability with existing benchmarks. We position our work at the intersection of three requirements: (1) rigorous cross-domain evaluation through unseen database splits, (2) comprehensive coverage of Chinese linguistic phenomena, and (3) practical support for enterprise SQL dialects. This tripartite focus enables more accurate assessment of model capabilities for real-world Chinese business intelligence applications, where all three dimensions must be addressed simultaneously.

\section{Corpus Construction}

Our benchmark combines carefully curated public datasets with enterprise-inspired synthetic cases to achieve both breadth and realism. The construction process follows five key phases:

\paragraph{Data Source Composition}
We integrate two data sources:

\begin{itemize}
    \item \textbf{Public Kaggle datasets}: 28 databases across 8 domains (finance, e-commerce, etc.) meeting strict criteria: (i) multi-table database must have verifiable primary-key and foreign-key (PK/FK) relationships, (ii) value diversity enabling meaningful filters, (iii) schema quality (readable names, proper typing).
    
    \item \textbf{Enterprise-inspired synthetic cases}: Using Ant Group's real user query patterns, we generated 120 additional cases (24\% of total) via: (a) template-based simulation of high-frequency business scenarios, (b) LLM-augmented paraphrasing of actual internal queries (anonymized and schema-adapted). These synthetic cases give Falcon realistic, enterprise-grade scenarios derived from real user patterns, enabling faithful simulation of production enterprise workflows while preserving privacy and ensuring compatibility with the MaxCompute/Hive dialect.
\end{itemize}

\paragraph{Annotation Protocol} 
The annotation protocol ensures each Chinese question is paired with a correct, executable MaxCompute/Hive SQL and enriched with fine-grained labels to support reproducibility, error analysis, and downstream model diagnostics. Bilingual experts (CS graduates with $\geq$2 years SQL experience) follow a strict workflow:
\begin{enumerate}
    \item \textit{Question authoring}: Compose Chinese questions using business vernacular (e.g. top 3 MoM-growing categories)
    \item \textit{SQL development}: Write executable MaxCompute/Hive queries following our style guide emphasizing:
    \begin{itemize}
        \item Structural clarity: Explicit JOIN conditions over implicit joins
        \item Dialect correctness: Proper handling of date functions, string collation
        \item Value safety: Parameterized filters when literal values appear
    \end{itemize}
    \item \textit{Cross-validation}: Independent execution by two engineers with result comparison
\end{enumerate}

\paragraph{Quality Assurance}
We implement a rigorous multi-stage verification process to ensure benchmark quality.

\textbf{Automated checks} cover: (1) SQL executability through engine validation, (2) result consistency via differential testing, and (3) enterprise compliance using a custom dialect linter that detects non-portable MaxCompute/Hive syntax. The validation pipeline rejects queries with implicit cross joins, redundant table scans, or non-deterministic constructs.

\textbf{Manual review} focuses on: (1) runtime error analysis for edge cases, (2) semantic equivalence verification through independent execution, and (3) business logic review by domain experts. For Chinese-specific aspects, we verify proper handling of ellipsis resolution (e.g., recovering omitted comparison operators) and business term mapping (e.g., "GMV" to actual column references).

This dual approach achieves: (1) 92\% inter-annotator agreement measured on 120 audit items, (2) 100\% syntax compliance with target dialects through iterative refinement, and (3) 85\% coverage of enterprise query patterns as measured against Ant Group's internal taxonomy. All gold SQL solutions are verified to return non-empty result sets unless explicitly modeling empty-result scenarios.

\paragraph{Innovative Features}
Compared to prior work, Falcon introduces:
\begin{itemize}
    \item \textbf{Hybrid data sourcing}: Combines public datasets' diversity with enterprise patterns' authenticity
    \item \textbf{Dialect-aware validation}: Specialized checks for MaxCompute/Hive edge cases
    \item \textbf{Business semantic tagging}: 15 categories of Chinese business expressions
\end{itemize}

This methodology ensures Falcon bridges the gap between academic benchmarks and real-world deployment requirements while maintaining rigorous quality standards.

\section{Dataset Statistics and Comparison}
\paragraph{Data Composition and Scale.}
The \benchmark corpus comprises 600 Chinese questions with dual provenance: (1) 500 questions (83.3\%) from 28 curated public Kaggle databases, and (2) 100 enterprise-inspired questions (16.7\%) synthesized from anonymized Ant Group query patterns. This hybrid composition balances broad domain coverage with authentic business scenario representation.

\begin{figure}[t]
\centering
\begin{tikzpicture}
\begin{axis}[
  ybar stacked,
  width=\linewidth,
  height=7.0cm,
  bar width=9pt,
  ymin=0,
  ymax=140,
  ylabel={Question Count},
  symbolic x coords={Finance,E-commerce,Retail,Sports,Transportation,Sci-Fi,Mobile,Education,IT,MacroEcon,SocioEcon},
  xtick=data,
  xticklabel style={font=\scriptsize, rotate=30, anchor=east},
  ymajorgrids,
  grid style={black!12},
  enlarge x limits=0.06,
  legend style={
    font=\scriptsize,
    cells={anchor=west},
    legend columns=2,
    at={(0.5,1.05)},
    anchor=north
  },
  nodes near coords,
  nodes near coords style={font=\tiny, color=black, yshift=2pt}
]
\addplot [fill=blue!60, draw=blue!70!black] coordinates {
  (Finance,82) (E-commerce,110) (Retail,70) (Sports,45) 
  (Transportation,25) (Sci-Fi,28) (Mobile,13) (Education,14) 
  (IT,7) (MacroEcon,13) (SocioEcon,8)
};
\addplot [fill=orange!60, draw=orange!70!black] coordinates {
  (Finance,15) (E-commerce,21) (Retail,5) (Sports,9)
  (Transportation,4) (Sci-Fi,6) (Mobile,3) (Education,2)
  (IT,1) (MacroEcon,2) (SocioEcon,2)
};
\legend{Kaggle Public Data (500), Ant Synthetic Data (100)}
\end{axis}
\end{tikzpicture}
\caption{Distribution showing benchmark composition (500 public + 100 enterprise-inspired questions). Ant-sourced cases concentrate in high-value domains like Finance (15/97) and E-commerce (21/131).}
\label{fig:domain}
\end{figure}
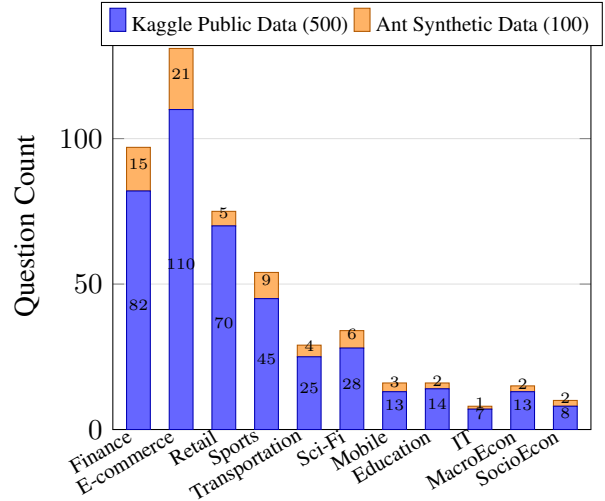

\paragraph{Structural Characteristics.} 
The corpus exhibits graded complexity:
\begin{itemize}[leftmargin=1.5em,noitemsep]
  \item \textbf{Schema scale}: 77\% multi-table questions with $\geq$4-table joins in 52\% cases
  \item \textbf{SQL operators}: Aggregation, CTEs, and window functions
\end{itemize}

\paragraph{Linguistic Profiling.}
The benchmark captures distinctive Chinese linguistic phenomena that challenge text-to-SQL systems:

\begin{itemize}[leftmargin=1.5em,noitemsep]
    \item \textbf{Ellipsis and anaphora}: Frequent omission of implied elements (e.g., "[compared to last year]" in "sales growth [compared to last year]")
    \item \textbf{Business terminology}: Domain-specific expressions like "MoM growth" (month-over-month) and "GMV top-10" 
    \item \textbf{Numeric expressions}: Mix of verbal ("twenty thousand") and numeric (20,000) forms
    \item \textbf{Modifier scope}: Ambiguous phrase attachments in complex noun phrases
\end{itemize}

Enterprise-inspired questions exhibit more frequent and complex instances of these phenomena compared to public dataset questions, particularly in nested business metric calculations, time-window comparisons, and relative performance evaluations.

\section{Task Definition and Evaluation Metrics}
\paragraph{Problem setting.}
Given a Chinese question $Q$, a database $D$ (tables, columns, types, PK/FK) and type-aware sampled values, generate MaxCompute/Hive SQL $S$ such that $V(S,D)$ is semantically equivalent to the gold answer $V^\ast$.

\paragraph{Inputs and outputs.}
Inputs and outputs. Input includes the question, Data Definition Language (DDL) statements with a compact PK/FK graph, and up to 8 sampled values per column; the output is an executable SQL query, for which Common Table Expressions (CTEs) are encouraged to handle multi-step logic.

\paragraph{Scope.}
Single-turn; no external knowledge; deterministic execution-based evaluation; dialect constraints enforced. Non-deterministic functions (e.g., rand) are disallowed.

\subsection{Comparator and Normalization}
To ensure reliable, name-agnostic matching under realistic variations in projection, ordering, and aliases, both predicted outputs and gold answers are normalized before comparison. We canonicalize types (including numeric precision/scale), normalize case and whitespace for strings, and standardize timestamps to a fixed timezone and format. Results are represented as column vectors stored in a column-oriented JSON format (keys are column names, values are full column arrays), but column names are ignored during matching. Unless the SQL explicitly contains an ORDER BY clause, row order is treated as immaterial and results are compared as multisets of rows; if ORDER BY is present, sequence equality is enforced.

\subsection{Content-based Result Set Evaluation}

To ensure a robust evaluation that is insensitive to cosmetic differences like column renaming, reordering, or redundant projections, we adopt a content-based evaluation method. This approach moves beyond simple text comparison to assess the semantic equivalence of query results at the column-vector level.

The process begins by verifying coverage. We generate a content-based MD5 hash for each normalized column in both the predicted and gold results. A prediction is considered to have sufficient coverage only if the multiset of its column hashes contains the entire multiset of the gold column hashes. This step tolerates extra columns in the prediction, as long as all required columns are present.

Following a successful coverage check, we perform column alignment to map each gold column to its corresponding predicted column using their identical content hashes. Ties, which occur if multiple predicted columns have the same content, are deterministically resolved by preferring the column with the closest data type signature, and then the one with the earliest index.

Once alignment is complete, any extra columns in the prediction are ignored, and the final row comparison is performed on the aligned projection only. The strictness of this comparison depends on the gold query: we require sequence equality (exact order) if an ORDER BY clause was used, and multiset equality (any order) otherwise. An empty gold result is deemed correct only if the prediction yields an empty set as well.

\section{Experiments and Analysis}
\label{sec:experiments}
\paragraph{Setup.}
We evaluate 13 Large Language Models (LLMs) using one-pass decoding with fixed settings (temperature 0.0, seed 3, max output 8192 tokens). Prompts include: (i) the Chinese question; (ii) the schema DDL with a PK/FK summary; (iii) up to 8 sampled values per column; and (iv) an explicit instruction to use the MaxCompute dialect and avoid non-deterministic functions. To encourage structured reasoning, we prompt models to use stepwise planning and Common Table Expressions (CTEs) for queries involving three or more joins.

Our primary evaluation metric is Exact Result Accuracy (ExAcc), which measures whether a model's predicted SQL query produces a result set that is semantically equivalent to the gold standard. This comparison is performed using the robust, content-based evaluation method described in Section~\ref{sec:content-eval}, which is tolerant to variations in column order, naming, and superfluous projections.

\subsection{Aggregate Results}
Across the full set, the best model (DeepSeek-R1) achieves 45.2\% \exacc, while the weakest (Qwen3-32B) reaches 20.2\%. No model exceeds 50\%, indicating substantial headroom under enterprise constraints. Reasoning-oriented models outperform instruction-only counterparts, especially on multi-join pipelines and nested filters.

\begin{table}[t]
  \centering
  \small
  \caption{Model accuracy on the Falcon benchmark (Exact Result Accuracy). All results are computed over 500 Chinese questions with MaxCompute-compatible SQL.}
  \label{tab:summary-acc}
  \begin{tabular}{@{}lS[table-format=2.1]@{}}
    \toprule
    \textbf{Model} & {\textbf{Accuracy (\%)}} \\
    \midrule
    DeepSeek-R1 \cite{deepseek-llm-2024} & \textbf{45.2} \\
    o1 \cite{yi-models-2024} & 43.0 \\
    o3-mini & 42.2 \\
    Claude-3.7-Sonnet-Thinking \cite{claude-3-7-2024} & 41.0 \\
    GPT-4.1 \cite{gpt4o-2024} & 40.2 \\
    Claude-3.7-Sonnet & 40.0 \\
    Qwen3-Coder-480B-a35B-Instruct \cite{qwen2-2024} & 37.0 \\
    Gemini 2.5 Pro \cite{gemini-1.5-2024} & 36.8 \\
    Gemini 2.5 Pro Reasoning & 34.8 \\
    Llama 3.3-70B \cite{llama3-2024} & 24.0 \\
    GPT-4o & 23.4 \\
    DeepSeek-R1-Distill-Qwen-32B & 23.4 \\
    Qwen3-32B & 20.2 \\
    \bottomrule
  \end{tabular}
  \caption*{\footnotesize Note: Accuracy is computed using a robust result comparator tolerant to projection/order/alias variations.}
\end{table}


\subsection{Stratified Results}
\label{subsec:stratified-falcon}

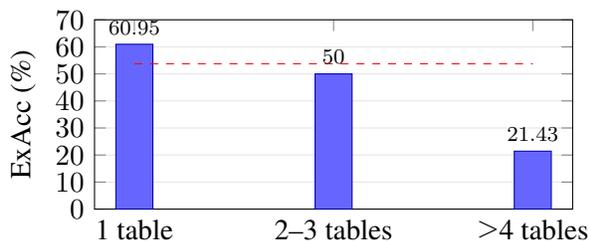
\begin{figure}[h!]
  \centering
  \begin{tikzpicture}
    \begin{axis}[
      width=\linewidth, height=4.1cm, ymin=0, ymax=70,
      ymajorgrids, grid style={black!10},
      symbolic x coords={1 table,2--3 tables,$\geq$4 tables},
      xtick=data, ylabel={ExAcc (\%)},
      nodes near coords, nodes near coords style={font=\scriptsize},
      bar width=14pt, ytick={0,10,20,30,40,50,60,70}
    ]
      \addplot[ybar, fill=blue!60, draw=blue!70!black]
        coordinates {(1 table,60.95) (2--3 tables,50.00) ($\geq$4 tables,21.43)};
      \draw[dashed, red] (axis cs:1 table,53.75) -- (axis cs:$\geq$4 tables,53.75);
      \node[red,anchor=west,font=\scriptsize] at (axis cs:$\geq$4 tables,54.6) {};
    \end{axis}
  \end{tikzpicture}
  \vspace{-6pt}
  \caption{ExAcc by join width. The corpus mean is indicated by the dashed line.}
  \label{fig:joinwidth-falcon}
\end{figure}

\paragraph{Join width.}
Accuracy decays monotonically with schema width (Fig.~\ref{fig:joinwidth-falcon}). Single-table queries reach \textbf{60.95\%} ExAcc, exceeding the corpus mean (\textbf{53.75\%}); 2–3 table queries show a moderate decline to \textbf{50.00\%}; queries with $\geq$4 tables drop to \textbf{21.43\%}. The sharp fall in the $\geq$4 bucket indicates two compounding factors: (i) fragile join-key propagation across longer chains (key threading through multiple hubs), and (ii) confusions among near-synonymous attributes (e.g., date vs.\ period; price vs.\ cost) that are amplified when multiple tables expose semantically adjacent fields.

\paragraph{Operator set.}
All items in this evaluation fall under the \emph{Nested/CTE} family (no aggregation, window, or set operations are present). The resulting ExAcc for the Nested/CTE slice equals the corpus mean (\textbf{53.75\%}). In this configuration, join width is the dominant driver of difficulty. The absence of window/set operations also avoids dialect-specific failure modes that typically affect those operators.

\subsection{Error Profile}
\label{subsec:error-falcon}
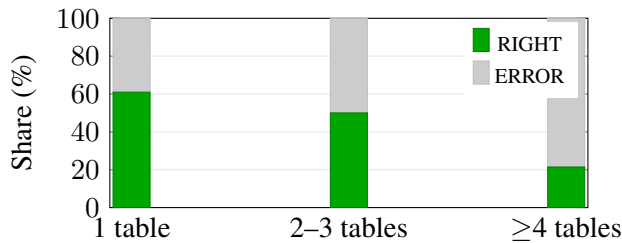
\begin{figure}[h!]
  \centering
  \begin{tikzpicture}
    \begin{axis}[
      width=\linewidth, height=4.1cm, ybar stacked, bar width=14pt,
      ymin=0, ymax=100, ytick={0,20,40,60,80,100},
      ylabel={Share (\%)}, ymajorgrids, grid style={black!10},
      symbolic x coords={1 table,2--3 tables,$\geq$4 tables}, xtick=data,
      legend style={at={(0.98,0.98)},anchor=north east,font=\scriptsize,draw=none},
      enlarge x limits=0.05
    ]
      \addplot[fill=green!65!black, draw=green!50!black] coordinates
        {(1 table,60.95) (2--3 tables,50.00) ($\geq$4 tables,21.43)};
      \addplot[fill=gray!40, draw=gray!50] coordinates
        {(1 table,39.05) (2--3 tables,50.00) ($\geq$4 tables,78.57)};
      \legend{RIGHT,ERROR}
    \end{axis}
  \end{tikzpicture}
  \vspace{-6pt}
  \caption{RIGHT/ERROR composition by join width. Error share escalates with schema width.}
  \label{fig:stacked-errors}
\end{figure}

To analyze failure modes, we categorize each prediction as either correct (\textbf{RIGHT}) if it achieves \exacc, or incorrect (\textbf{ERROR}) otherwise. Figure~\ref{fig:stacked-errors} summarizes the proportion of these outcomes across different join widths. The analysis reveals that the error mass is heavily concentrated in queries with $\geq$4 tables, where the error rate reaches 78.57\%. This indicates that schema linking and multi-hop key propagation are the principal bottlenecks for current models. This pattern is characteristic of enterprise schemas, where wide joins expose numerous semantically adjacent columns and create long dependency chains that challenge model reasoning.

\section{Conclusion}
\benchmark provides an enterprise-grounded, Chinese-focused benchmark with dual-axis annotations and reproducible execution-based evaluation. Results (20.2–45.2\% \exacc) highlight persistent challenges in wide-schema linking and semantics-to-operator alignment. We advocate relation-aware schema encoding~\cite{wang-etal-2020-ratsql}, lightweight preprocessing for Chinese ellipsis/coreference/business shorthand, and dialect-aware constrained decoding~\cite{scholak-etal-2021-picard}.

\paragraph{Ethics and availability.}
\benchmark uses public Kaggle datasets and de-identified question patterns. We release the benchmark specification, toolchain, container image, and documentation; any enterprise logs are anonymized.

\paragraph{Acknowledgements}
We thank the Kaggle community for publicly available datasets, and Ant Group engineers for anonymized schema patterns and question templates.


\appendix
\section*{Appendix}
\label{app:datasets}
\setcounter{table}{0}
\renewcommand{\thetable}{A\arabic{table}}

\paragraph{External datasets.}
Table~\ref{tab:external-datasets} enumerates the 28 external datasets referenced in Section~\ref{sec:data}. Each entry includes a stable Kaggle URL. For single-page, two-column proceedings, the table is typeset with small font and automatic line breaks. Long URLs will wrap; compilation requires \texttt{url}/\texttt{hyperref}.

\begin{table*}[t]
  \centering
  \scriptsize
  \setlength{\tabcolsep}{6pt}
  \begin{tabular}{@{}r p{0.86\linewidth}@{}}
    \toprule
    \textbf{ID} & \textbf{Source (Kaggle URL)} \\
    \midrule
    1  & \url{https://www.kaggle.com/datasets/nitindatta/finance-data?select=Finance_data.csv} \\
    2  & \url{https://www.kaggle.com/datasets/krishnaraj30/finance-loan-approval-prediction-data?select=test.csv} \\
    3  & \url{https://www.kaggle.com/datasets/iveeaten3223times/massive-yahoo-finance-dataset?resource=download} \\
    4  & \url{https://www.kaggle.com/datasets/hhenry/finance-factoring-ibm-late-payment-histories/data} \\
    5  & \url{https://www.kaggle.com/datasets/ramjasmaurya/unicorn-startups} \\
    6  & \url{https://www.kaggle.com/datasets/shantanugarg274/sales-dataset} \\
    7  & \url{https://www.kaggle.com/datasets/gregorut/videogamesales} \\
    8  & \url{https://www.kaggle.com/datasets/lava18/google-play-store-apps} \\
    9  & \url{https://www.kaggle.com/datasets/ashishraut64/internet-users} \\
    10 & \url{https://www.kaggle.com/datasets/mattiuzc/stock-exchange-data?select=indexInfo.csv} \\
    11 & \url{https://www.kaggle.com/datasets/hosubjeong/bakery-sales} \\
    12 & \url{https://www.kaggle.com/datasets/mexwell/google-merchandise-sales-data} \\
    13 & \url{https://www.kaggle.com/datasets/aslanahmedov/walmart-sales-forecast} \\
    14 & \url{https://www.kaggle.com/datasets/mysarahmadbhat/toy-sales?select=products.csv} \\
    15 & \url{https://www.kaggle.com/datasets/ananta/credit-card-data/data} \\
    16 & \url{https://www.kaggle.com/datasets/bernardnm/great-school?select=Teachers.csv} \\
    17 & \url{https://www.kaggle.com/datasets/bytadit/ecommerce-order-dataset} \\
    18 & \url{https://www.kaggle.com/datasets/madhurpant/world-economic-data?select=cost_of_living.csv} \\
    19 & \url{https://www.kaggle.com/datasets/thedevastator/relationship-between-alcohol-consumption-and-lif?select=drinks.csv} \\
    20 & \url{https://www.kaggle.com/datasets/rishabhrajsharma/cityride-dataset-rides-data-drivers-data?select=Rides_Data.csv} \\
    21 & \url{https://www.kaggle.com/datasets/aymenberkani/data-cleaning-for-a-customer-database?select=e_orders.csv} \\
    22 & \url{https://www.kaggle.com/datasets/frankienoguera/2024-25-ufc-relational-database?select=ufc_events_stats.csv} \\
    23 & \url{https://www.kaggle.com/datasets/atharvasoundankar/ben-10-alien-universe-realistic-battle-dataset?select=ben10_aliens.csv} \\
    24 & \url{https://www.kaggle.com/datasets/akxiit/blinkit-sales-dataset} \\
    25 & \url{https://www.kaggle.com/datasets/farhadzeynalli/techsales-bakutech} \\
    26 & \url{https://www.kaggle.com/datasets/technika148/football-database} \\
    27 & \url{https://www.kaggle.com/datasets/andrexibiza/grocery-sales-dataset} \\
    28 & \url{https://www.kaggle.com/datasets/marthadimgba/online-shop-2024} \\
    \bottomrule
  \end{tabular}
  \caption{External datasets used or referenced in experiments. URLs are provided for reproducibility.}
  \label{tab:external-datasets}
\end{table*}


\end{document}